\documentclass{article}

\usepackage{silence}
\usepackage[preprint]{neurips_2025}
\usepackage[utf8]{inputenc} % allow utf-8 input
\usepackage[T1]{fontenc}    % use 8-bit T1 fonts
\usepackage{hyperref}       % hyperlinks
\usepackage{url}            % simple URL typesetting
\usepackage{booktabs}       % professional-quality tables
\usepackage{amsfonts}       % blackboard math symbols
\usepackage{nicefrac}       % compact symbols for 1/2, etc.
\usepackage{microtype}      % microtypography
\usepackage{graphicx}       % to include images
\usepackage{amsmath}        % for equations
\usepackage{subcaption}     % for subfigures
\usepackage{xcolor}         % for colors
\usepackage{subcaption}
\usepackage[percent]{overpic}
\usepackage{tikz}

  \tikzset{titlebox/.style={
    draw=gray!50, 
    fill=gray!5, 
    rounded corners=2pt, 
    inner sep=3pt, 
    font=\small
  }}

  \tikzset{titlebox2/.style={
    draw=gray!50, 
    fill=gray!5, 
    rounded corners=2pt, 
    inner sep=3pt
  }}

\title{Training-Free Diffusion Priors for Text-to-Image Generation via Optimization-based Visual Inversion}

\author{
  Samuele Dell'Erba\\
  University of Florence\\
  \texttt{samuele.dellerba@edu.unifi.it} \\
  \And
  Andrew D. Bagdanov\\
  University of Florence\\
  \texttt{andrew.bagdanov@unifi.it} \\
}

\begin{document}

\maketitle

%======================================================================
\begin{abstract}
Diffusion models have established the state-of-the-art in text-to-image generation, but their performance often relies on a diffusion prior network to translate text embeddings into the visual manifold for easier decoding. These priors are computationally expensive and require extensive training on massive datasets. In this work, we challenge the necessity of a trained prior at all by employing Optimization-based Visual Inversion (OVI), a training-free and zero-shot alternative, to replace the need for a prior. OVI initializes a latent visual representation from random pseudo-tokens and iteratively optimizes it to maximize the cosine similarity with the input textual prompt embedding. We further propose two novel constraints, a Mahalanobis-based and a Nearest-Neighbor loss, to regularize the OVI optimization process toward the distribution of realistic images. Our experiments, conducted on Kandinsky 2.2, show that OVI can serve as an alternative to traditional priors. More importantly, our analysis reveals a critical flaw in current evaluation benchmarks like T2I-CompBench++, where simply using the text embedding as a prior achieves surprisingly high scores, despite lower perceptual quality. Our constrained OVI methods improve visual fidelity over this baseline, with the Nearest-Neighbor approach proving particularly effective. It achieves quantitative scores comparable to or higher than the state-of-the-art data-efficient prior, underscoring the potential of optimization-based strategies as viable, training-free alternatives to traditional priors. The code will be publicly available upon acceptance.
\end{abstract}

%======================================================================
\section{Introduction}

Text-to-image (T2I) generation has seen remarkable progress, evolving from autoregressive approaches like DALL-E~\cite{ramesh2021zero} to diffusion models~\cite{ho2020denoising, StableDiffusion}, which now represent the state-of-the-art. Architectures like unCLIP~\cite{unCLIP} have demonstrated that a hierarchical approach, which first generates a CLIP image embedding~\cite{radford2021learning} from a text caption using a prior model and then generates an image from that embedding using a decoder, can produce high-fidelity and diverse images. However, this performance comes at a significant cost: the diffusion prior is typically a large model that is expensive to train, requiring billions of image-text pairs and substantial computational resources.

Recent work has focused on making these priors more efficient. ECLIPSE~\cite{Eclipse}, for example, introduced a data-efficient prior that achieves competitive results at a fraction of the parameters and training samples. Despite these advancements, the fundamental paradigm remains unchanged: the prior must still be learned through an extensive training phase, limiting accessibility and adaptability.

In this work, we investigate a radical alternative: ``can we entirely eliminate the training phase for the prior?'' Building on recent work on modality inversion~\cite{CrossTheGap}, we leverage Optimization-based Visual Inversion (OVI), a training-free method proposed therein to cast intra-modal tasks into inter-modal ones, to substitute the prior with an on-the-fly optimization process. Instead of relying on a mapping network, OVI generates a CLIP image embedding by iteratively refining a random latent vector to maximize its semantic alignment with the target textual prompt embedding. To further guide this process towards plausible visual representations, we introduce two novel regularization strategies: a Mahalanobis constraint~\cite{lee2018simple} that pulls the embedding towards the global distribution of real image embeddings, and a Nearest-Neighbor constraint, inspired by retrieval-augmented approaches~\cite{blattmann2022retrieval, chen2022re, sheynin2022knn}, that aligns it with the most similar real image embedding from a reference dataset.

Our preliminary results reveal several interesting aspects of T2I generation. We find that unconstrained OVI produces qualitatively similar results to directly using the text embedding in the decoder, a surprisingly strong baseline first noted in~\cite{unCLIP}. This baseline achieves state-of-the-art scores on the benchmark T2I-CompBench++~\cite{T2ICompBench}, often exceeding expensive trained priors like ECLIPSE. This suggests a potential ``metric problem,'' consistent with the misalignment of metrics like FID~\cite{heusel2017gans} and CLIPScore~\cite{hessel2021clipscore} with human preference~\cite{pickscore}. Specifically, current benchmarks seem to over-reward semantic similarity at the expense of visual realism. Our constrained OVI methods improve visual fidelity over this baseline, with the Nearest-Neighbor approach proving particularly effective. It achieves quantitative scores comparable to or higher than the state-of-the-art prior ECLIPSE while utilizing a reference set of only 40k images ($\approx$0.8\% of the 5 million used by the latter), highlighting the potential of optimization-based strategies as resource-efficient, training-free alternatives to data-hungry learned priors.

% The rest of this paper is structured as follows. Section 2 reviews related work. Section 3 details our proposed OVI methods. Section 4 presents our experimental results, and Section 5 discusses our findings and limitations before we conclude in Section 6.

%======================================================================
\section{Related Work}

In this section we review work from the recent literature most relevant to our contribution.

\subsection{Diffusion Models for Text-to-Image Generation}
Diffusion models have become the dominant architecture for T2I synthesis. The unCLIP~\cite{unCLIP} framework popularized a two-stage pipeline consisting of a prior and a decoder. The decoder, often a U-Net based model, is trained to generate an image conditioned on a CLIP image embedding. Our work leverages the publicly available Kandinsky 2.2 decoder~\cite{Kandinsky}, which is known for its high-quality image generation capabilities. While recent adapters like ControlNet~\cite{zhang2023adding} enable fine-grained structural control, our work focuses on the semantic guidance provided by the global image prior.

\subsection{Image Priors for Text-to-Image Generation}
The prior model is responsible for mapping the textual prompt embedding $z_y$ to the corresponding image embedding $z_x$. The original unCLIP~\cite{unCLIP} explored both autoregressive and diffusion-based priors, with the latter generally proving more efficient and effective. More recently, ECLIPSE~\cite{Eclipse} proposed a lightweight, non-diffusion prior trained with a combination of a projection and a contrastive loss, significantly reducing computational requirements while maintaining high performance. Our work investigates the possibility of omitting the prior training phase, proposing a zero-shot, optimization-based substitute.

Interestingly, the fundamental role of the prior has been partially challenged by the observation that feeding the CLIP text embedding directly into the decoder can produce surprisingly reasonable results. This phenomenon, first noted in the original unCLIP paper~\cite{unCLIP}, suggests that the image decoder possesses some inherent capability to interpret text embeddings, even if they do not lie on the natural image manifold. This observation establishes a strong, albeit lower-quality, baseline for any prior-free method and motivates the search for techniques that can improve upon this simple substitution without requiring a fully trained prior.

\subsection{Modality Inversion and Optimization-based Methods}
The idea of inverting a model or optimizing in a latent space is not new. The concept of test-time optimization has roots in the seminal Deep Image Prior~\cite{ulyanov2018deep}, while more recently, Null-text Inversion~\cite{mokady2023null} applied pivotal tuning to the unconditional embedding for editing real images. In the context of CLIP,~\cite{CrossTheGap} investigated modality inversion by adapting Textual Inversion (OTI)~\cite{gal2022image} and introducing Optimization-based Visual Inversion (OVI) to expose and bridge the gap between the text and image embedding manifolds~\cite{liang2022mind}. In parallel, methods like SEARLE~\cite{SEARLE} have used optimization to project a reference image into a ``pseudo-token'' for composed image retrieval, demonstrating the power of training-free optimization to achieve strong multimodal alignment. Our OVI-based approach builds on these concepts, adapting modality inversion to serve as a generative, training-free prior for text-to-image generation.

%======================================================================
\section{Our approach: Optimization-based Image Priors}

In this section we describe our optimization-based approach for deriving image priors from a pre-trained CLIP model. 

\subsection{Preliminaries}
A typical hierarchical T2I pipeline consists of a text encoder $E_T$, an image encoder $E_I$, a prior $g_\phi$, and a decoder $h_\theta$. Both the prior and the decoder are typically trained separately.

The diffusion decoder $h_\theta$ is trained to predict the noise $\epsilon$ added to an image $x$ at a timestep $t$, conditioned on the image embedding $z_x$ and optionally the text embedding $z_y$. Its loss function is:
\begin{equation}
    \mathcal{L}_{decoder} = \mathbb{E}_{\epsilon \sim \mathcal{N}(0, I), t, (z_x, z_y)} \left[ || \epsilon - h_\theta(x^{(t)}, t, z_x, z_y) ||_2^2 \right]
\end{equation}

The standard diffusion prior $g_\phi$ is trained to predict the ``clean'' image embedding $z_x$ from a noised version $z_x^{(t)}$, conditioned on the text embedding $z_y$:
\begin{equation}
    \mathcal{L}_{prior} = \mathbb{E}_{t, z_x^{(t)}} \left[ || z_x - g_\phi(z_x^{(t)}, t, z_y) ||_2^2 \right]
\end{equation}

More efficient priors like ECLIPSE~\cite{Eclipse} replace the diffusion model with a lightweight network $g_\phi$ trained with a combined loss. The goal is to make the predicted image embedding $\hat{z}_x = g_\phi(\epsilon, z_y)$ align with the ground-truth $z_x$. This is achieved with a projection objective:
\begin{equation}
    \mathcal{L}_{proj} = \mathbb{E}_{\epsilon \sim \mathcal{N}(0, I), (z_y, z_x)} \left[ || z_x - g_\phi(\epsilon, z_y) ||_2^2 \right]
\end{equation}
To further enforce semantic alignment, a CLIP contrastive loss is added, which encourages the predicted embedding $\hat{z}_x$ to be closer to its corresponding text embedding $z_y$ than to other text embeddings in the batch:
\begin{equation}
    \mathcal{L}_{CLS, y \to x} = -\frac{1}{N} \sum_{i=0}^{N} \log \frac{\exp(\text{sim}(\hat{z}_x^i, z_y^i) / \tau)}{\sum_{j \in [N]} \exp(\text{sim}(\hat{z}_x^i, z_y^j) / \tau)}
\end{equation}
where $\tau$ is a temperature parameter and $\text{sim}(\cdot, \cdot)$ denotes cosine similarity. The final objective is:
\begin{equation}
    \mathcal{L}_{ECLIPSE} = \mathcal{L}_{proj} + \lambda \cdot \mathcal{L}_{CLS, y \to x}
\end{equation}
Our work replaces the trained prior $g_\phi$ entirely with an optimization procedure. We use pre-trained CLIP encoders and the pre-trained Kandinsky 2.2 decoder, keeping them frozen.

\subsection{Optimization-based Visual Inversion (OVI)}
The optimization-based approach frames the prior's task as an optimization problem. Instead of learning a mapping, we directly search for an image embedding $\hat{z}_x$ that aligns with the given text embedding $z_y$. We initialize a set of $N$ learnable pseudo-tokens, which form our latent image embedding $z_{img}^{(0)}$. We then iteratively update this embedding for $T$ steps using an optimizer to minimize a loss function based on cosine similarity with the target text embedding, as computed by the CLIP encoders:
\begin{equation}
    L_{OVI} = 1 - \text{sim}(z_{img}^{(t)}, z_y)
\end{equation}
At each step $t$, the embedding $z_{img}^{(t)}$ is updated using the AdamW optimizer~\cite{AdamW} to reduce this loss. The final optimized embedding, $z_{img}^{(T)}$, serves as the input $\hat{z}_x$ to the decoder.

\subsection{Mahalanobis-constrained Visual Inversion}
While OVI maximizes semantic alignment, it does not guarantee that the resulting embedding $\hat{z}_x$ lies on the manifold of realistic image embeddings. To address this, we introduce a regularization term based on the Mahalanobis distance. We pre-compute the mean $\mu_{coco}$ and the inverse covariance matrix $\Sigma^{-1}_{coco}$ of image embeddings from a large reference dataset (MS-COCO~\cite{lin2014microsoft}). The Mahalanobis loss penalizes embeddings that are far from the center of this distribution:
\begin{equation}
    L_{Mahalanobis} = \sqrt{(z_{img}^{(t)} - \mu_{coco})^T \Sigma^{-1}_{coco} (z_{img}^{(t)} - \mu_{coco})}
\end{equation}
The total loss becomes a weighted sum:
\begin{equation}
    L_{Total} = L_{OVI} + \lambda_M \cdot L_{Mahalanobis}
\end{equation}
where $\lambda_M$ is a hyperparameter balancing semantic alignment and distributional realism.

\subsection{Nearest-Neighbor Constrained Visual Inversion}
As an alternative to a global distributional constraint, we propose a more local, instance-based regularizer. For a given text prompt $y$, we first find its nearest neighbor in the image embedding space from our reference dataset. Specifically, we find the image embedding $z_{closest}$ from the COCO dataset that has the highest cosine similarity with the text embedding $z_y$. We then add a loss term that encourages our optimized embedding $\hat{z}_x$ to be similar to this closest real embedding:
\begin{equation}
    L_{neighbor} = 1 - \text{cos\_sim}(z_{img}^{(t)}, z_{closest})
\end{equation}
The total loss is again a weighted sum:
\begin{equation}
    L_{Total} = L_{OVI} + \lambda_N \cdot L_{neighbor}
\end{equation}
This approach guides the optimization towards a specific, relevant region of the image manifold, rather than just its center.

%======================================================================
\section{Experimental Results}

In this section we report on a number of preliminary experiments performed to probe the effectiveness of our approach.

\subsection{Experimental Setup}

\textbf{Implementation Details.} 
We use the OpenCLIP~\cite{cherti2023reproducible} implementation of the ViT-bigG-14 model (trained on LAION-5B~\cite{schuhmann2022laion}) for text and image encoders and the Kandinsky 2.2~\cite{Kandinsky} model as our image decoder. All models are frozen during the OVI process.
Our primary baseline is ECLIPSE~\cite{Eclipse}, a state-of-the-art resource-efficient prior. We also compare against a strong baseline where the Direct Text Embedding ($z_y$) is fed directly to the decoder, which we refer to as \texttt{TextEmb}.
We use the MS-COCO (40k images) dataset~\cite{lin2014microsoft} to compute statistics for our constrained OVI methods. 

\textbf{Evaluation Setting.} We evaluate all methods on the T2I-CompBench++~\cite{T2ICompBench} benchmark, which measures compositional abilities across categories like color, shape, texture, and spatial relations. Unless otherwise specified, all our OVI methods use 1000 optimization steps, 6 pseudo-tokens, and generate 512x512 images with 50 diffusion steps (DDIM~\cite{song2021denoising}) using Classifier-Free Guidance~\cite{ho2022classifier} with a scale of 4.0 in the decoder. A fixed random seed is used for the decoder to ensure fair comparisons.

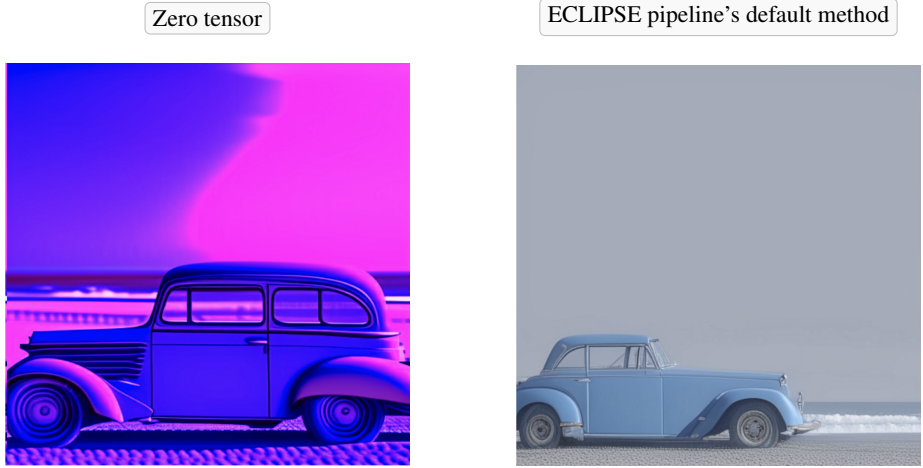
\begin{figure}[]
  \centering

  % --- Prima immagine (Sinistra) ---
  \begin{subfigure}{0.48\textwidth}
    \centering
    \begin{tikzpicture}
      \node[titlebox] {Zero tensor};
    \end{tikzpicture}
    
    \vspace{10pt} % <--- 1. AUMENTA lo spazio tra Titolo e Immagine
    
    \includegraphics[width=0.8\textwidth]{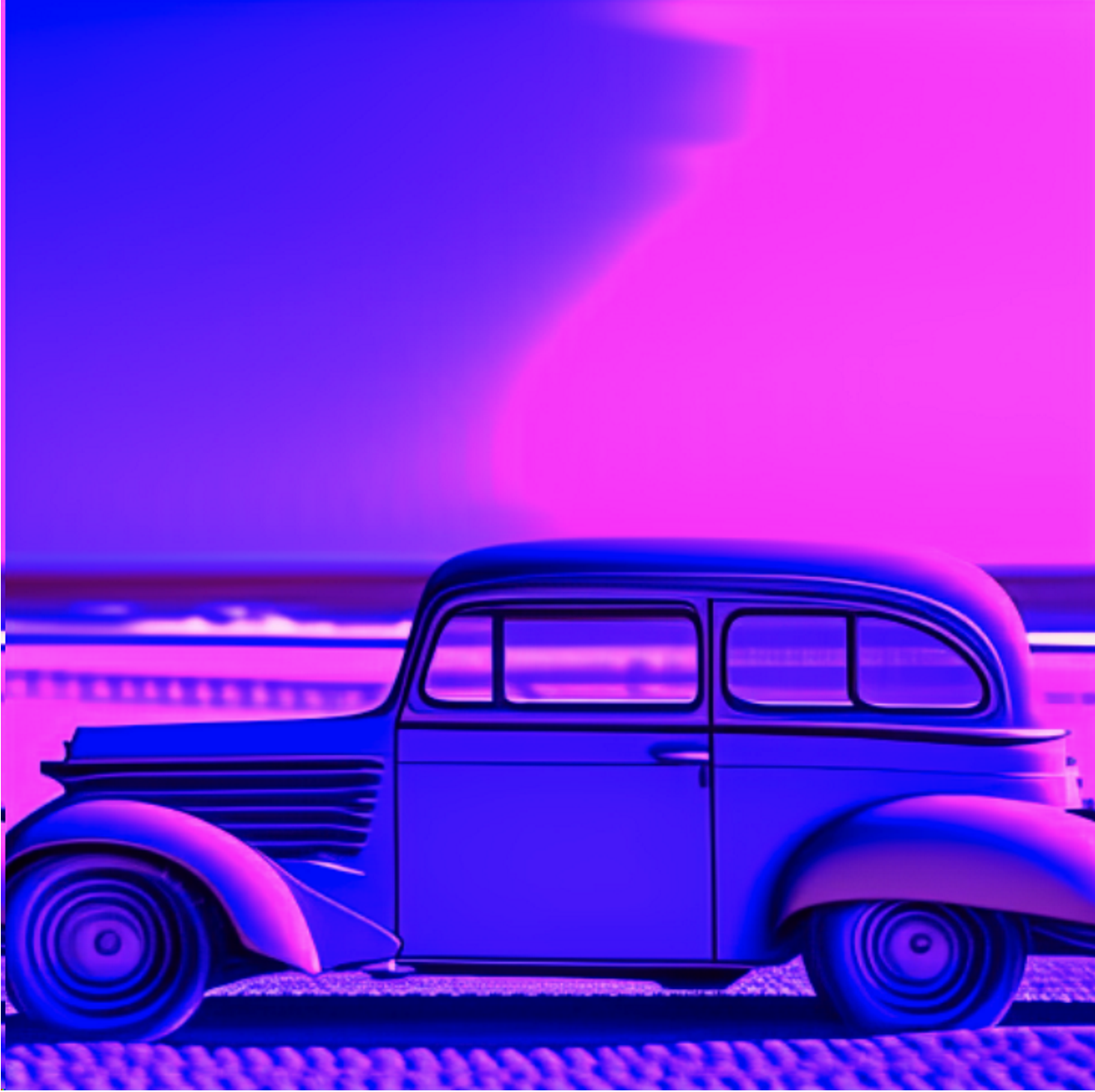}
    \end{subfigure}
  % --- Seconda immagine (Destra) ---
  \begin{subfigure}{0.48\textwidth}
    \centering
    \begin{tikzpicture}
      \node[titlebox] {ECLIPSE pipeline’s default method};
    \end{tikzpicture}
    
    \vspace{10pt} % <--- 1. AUMENTA lo spazio tra Titolo e Immagine
    
    \includegraphics[width=0.8\textwidth]{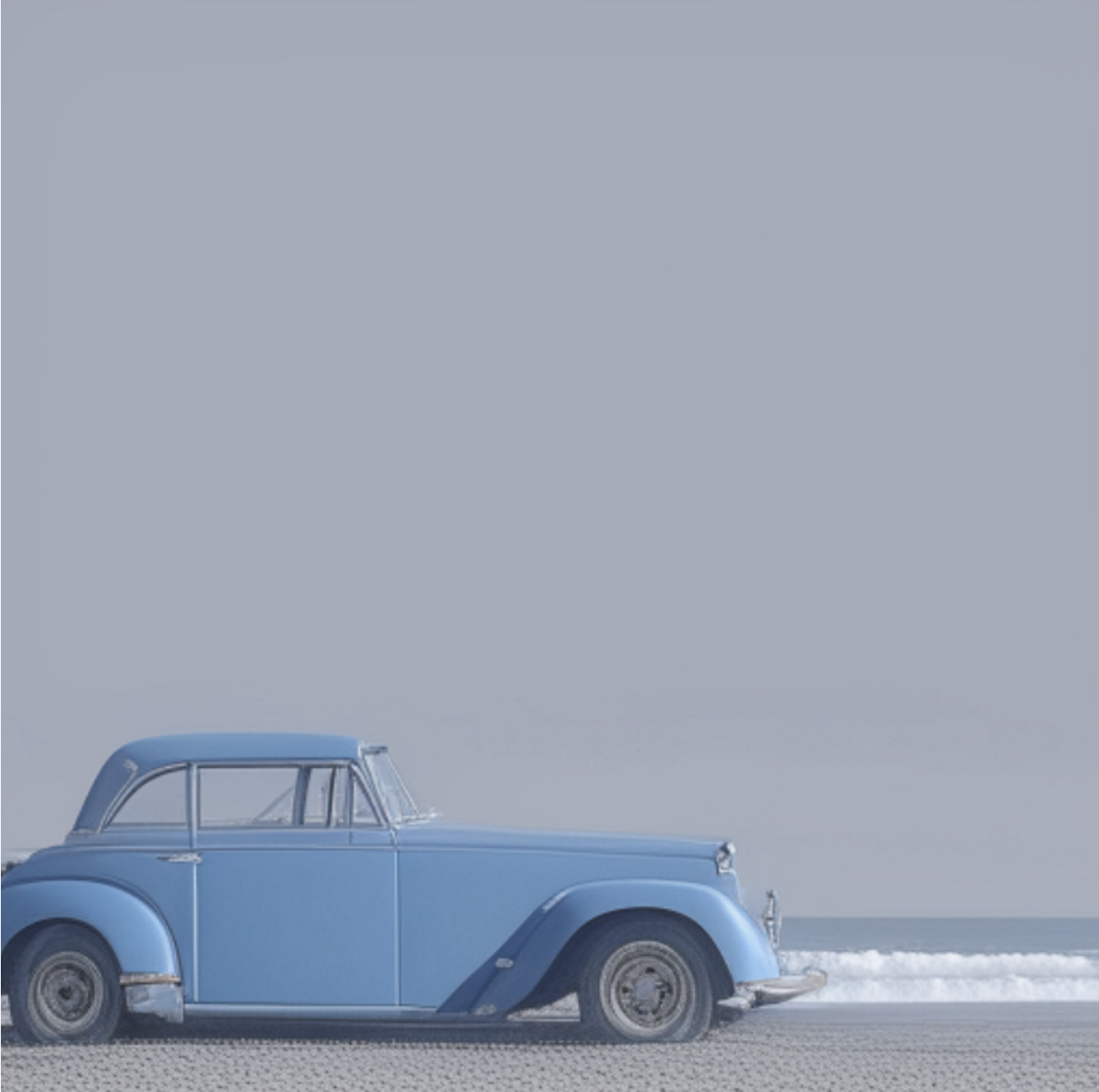}
   \end{subfigure}

  \caption{\small Impact of negative embedding initialization. We show results for the prompt ``blue old car on a beach'' (500 OVI steps, 1 Token, 512x512). \textbf{Left:} Using a zero tensor results in a strong violet filter artifact and poor quality. \textbf{Right:} Using the ECLIPSE pipeline's default method corrects the color balance and improves overall definition.}
  \label{fig:negative_embedding}
\end{figure}

\textbf{Additional Considerations.} Regarding the negative image embedding, our initial experiments using a zero-initialized tensor resulted in images with a pervasive violet color cast and degraded perceptual quality. We resolved this by adopting the default negative embedding generation method provided by the ECLIPSE pipeline, which effectively eliminated the color artifacts and significantly improved visual fidelity, as demonstrated in Figure \ref{fig:negative_embedding}. All experiments were run on a single NVIDIA RTX 4090 GPU.

\textbf{Evaluation Metrics.} We evaluate all methods on the T2I-CompBench++~\cite{T2ICompBench} benchmark. This benchmark provides a fine-grained assessment of compositional abilities. It does not rely on a single score but uses different models for different tasks:
\begin{itemize}
    \item \textbf{Attribute Binding (Color, Shape, Texture):} Evaluated using a Disentangled BLIP-VQA model~\cite{li2022blip}, which asks specific questions about object attributes (e.g., ``Is the book red?'').
    \item \textbf{2D Spatial Relations:} Evaluated using UniDet~\cite{zhou2022simple} to detect objects and verify the geometric relationship between their bounding box centers.
    \item \textbf{3D Spatial Relations:} Evaluated using UniDet combined with the DPT~\cite{ranftl2021vision} depth estimation model to check for occlusion and relative depth.
    \item \textbf{Non-spatial Relations:} Evaluated using CLIPScore~\cite{hessel2021clipscore}, which measures the cosine similarity between the prompt's text embedding and the generated image's embedding.
\end{itemize}
The final score is an average of the scores across these categories.

\subsection{Analysis of Unconstrained OVI}
Before presenting our main comparative results, we analyze the behavior of the unconstrained OVI method. We found that performance is highly dependent on key hyperparameters. Image quality at 256x256 resolution with a single pseudo-token is very low, worse even than the \texttt{TextEmb} baseline. Increasing the token count to 6 improves quality, but a substantial leap in fidelity is achieved at 512x512 resolution, where a good-quality image is formed within the first 100-200 optimization steps. We also observed that increasing the number of pseudo-tokens leads to more realistic images where the main subject tends to occupy a larger portion of the frame. Based on these preliminary experiments, we proceed with 512x512 resolution and 6 pseudo-tokens as a strong baseline for OVI.

\begin{figure}[]
  \centering
  \includegraphics[width=1\textwidth]{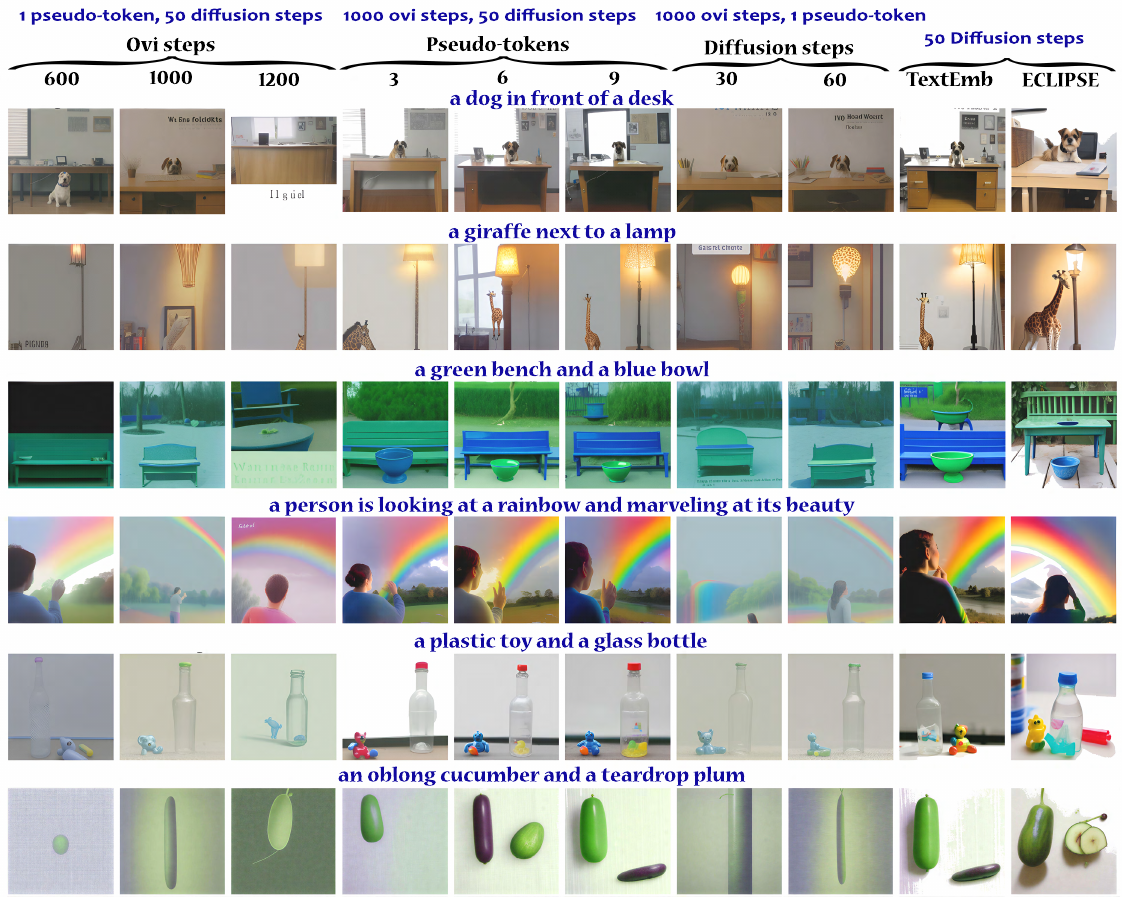}
  \caption{\small Qualitative comparison of OVI variants, the \texttt{TextEmb} baseline, and ECLIPSE~\cite{Eclipse}. Unconstrained OVI with more tokens produces results visually similar to \texttt{TextEmb}, while ECLIPSE yields higher fidelity.}
  \label{fig:ovi_vs_textemb}
\end{figure}

Qualitatively, we observe that the images generated by unconstrained OVI are semantically aligned with the prompt but often lack the photorealism of those produced by trained priors. As shown in Figure \ref{fig:ovi_vs_textemb}, the results with higher pseudo-tokens are visually very similar to the \texttt{TextEmb} baseline. This suggests that without constraints, OVI finds a point in the latent space that is semantically almost equivalent to the text embedding itself.

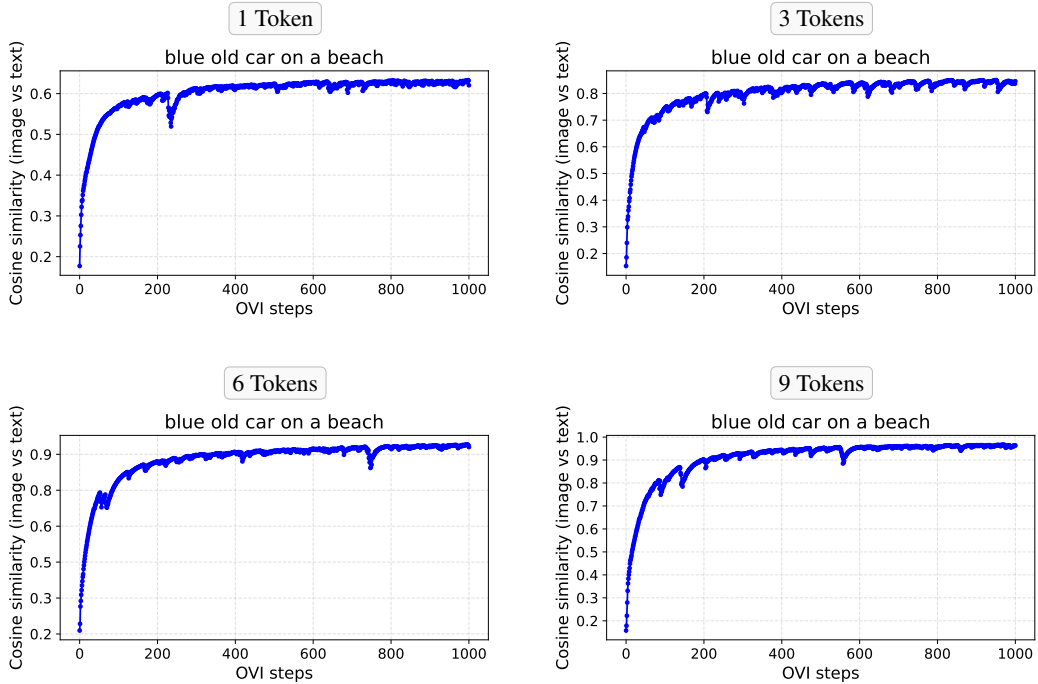
\begin{figure}[]
  \centering

  % --- Prima riga ---
  \begin{subfigure}{0.48\textwidth}
    \centering
    \begin{tikzpicture}
    \hspace{10pt}
      \node[titlebox] {1 Token};
    \end{tikzpicture}
    \vspace{2pt}
    \includegraphics[width=\textwidth]{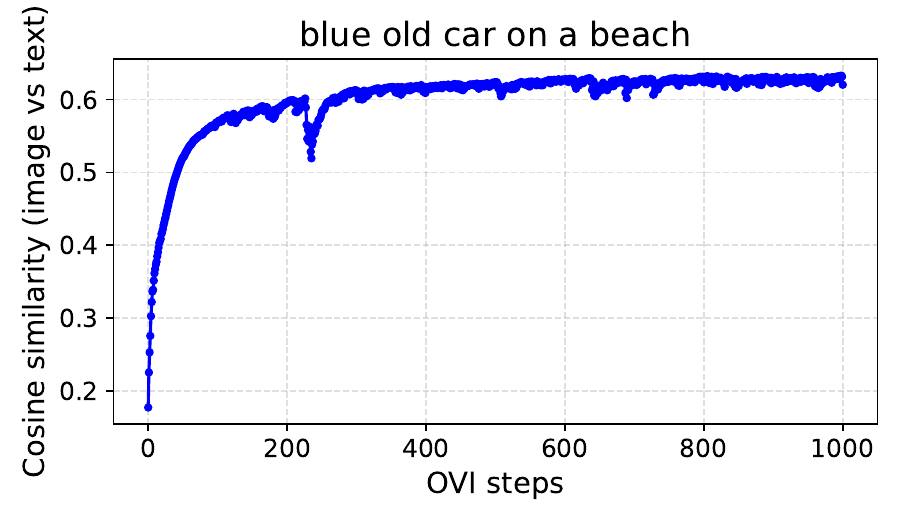}
  \end{subfigure}
  \hfill
  \begin{subfigure}{0.48\textwidth}
    \centering
    \begin{tikzpicture}
    \hspace{10pt}
      \node[titlebox] {3 Tokens};
    \end{tikzpicture}
    \vspace{2pt}
    \includegraphics[width=\textwidth]{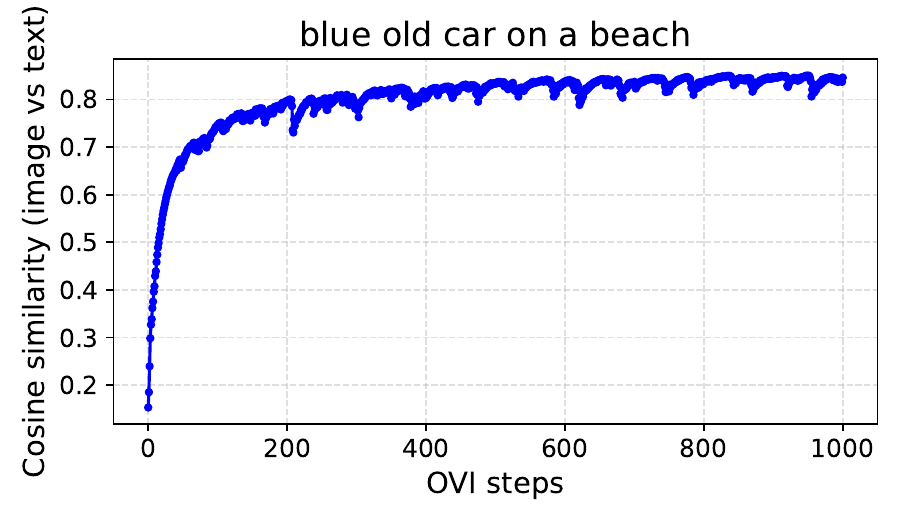}
  \end{subfigure}

  \vspace{1em} % Spazio verticale tra le due righe di grafici

  % --- Seconda riga ---
  \begin{subfigure}{0.48\textwidth}
    \centering
    \begin{tikzpicture}
    \hspace{10pt}
      \node[titlebox] {6 Tokens};
    \end{tikzpicture}
    \vspace{2pt}
    \includegraphics[width=\textwidth]{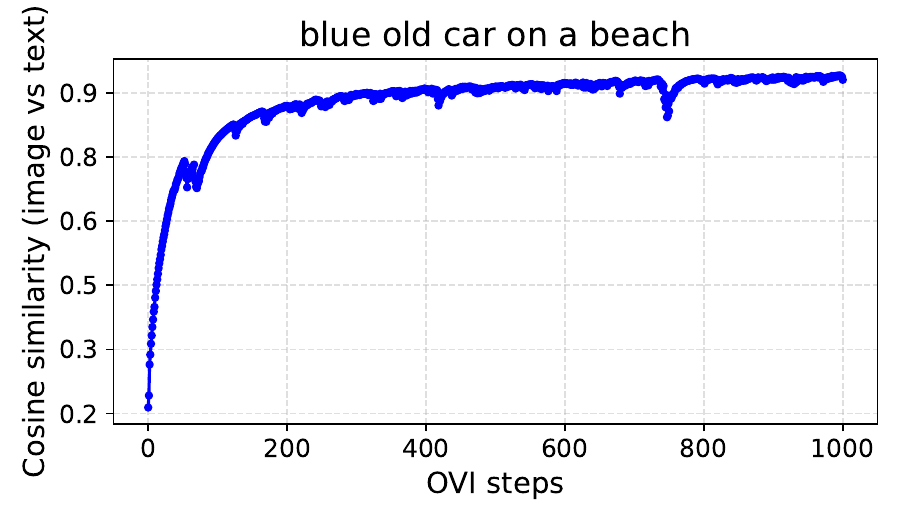}
  \end{subfigure}
  \hfill
  \begin{subfigure}{0.48\textwidth}
    \centering
    \begin{tikzpicture}
    \hspace{10pt}
      \node[titlebox] {9 Tokens};
    \end{tikzpicture}
    \vspace{2pt}
    \includegraphics[width=\textwidth]{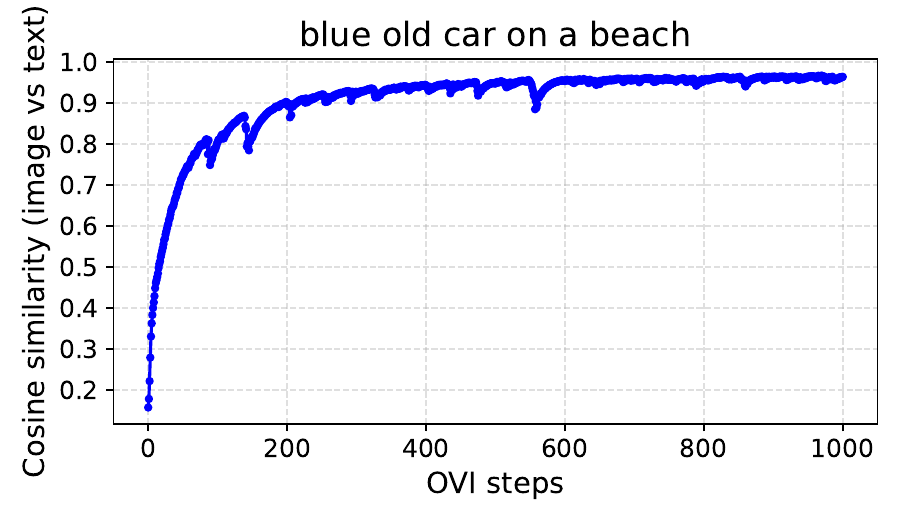}
  \end{subfigure}

  \caption{\small Cosine similarity between the optimized image embedding and the target text embedding over 1000 OVI steps. As the number of pseudo-tokens increases, the optimization reaches a higher similarity plateau, moving closer to the text embedding.}
  \label{fig:cosine_similarity}
\end{figure}

This convergence towards the text embedding is quantitatively confirmed by analyzing the cosine similarity, as visualized in Figure \ref{fig:cosine_similarity}. The image embedding produced by ECLIPSE consistently maintains a cosine similarity of approximately 0.65 with the text prompt. In contrast, for our unconstrained OVI, the similarity increases logarithmically with the number of optimization steps, eventually plateauing at a value determined by the number of pseudo-tokens. With 1000 steps, we reach a similarity of $\approx$0.63 with 1 token, $\approx$0.84 with 3 tokens, $\approx$0.93 with 6 tokens, and $\approx$0.96 with 9 tokens. This confirms our hypothesis: the optimization process is effectively pushing the image embedding out of the visual manifold and towards the text manifold, eventually converging to the text embedding itself. This explains why its performance on benchmarks mirrors that of the \texttt{TextEmb} baseline.

The benchmark scores presented in Table \ref{tab:ablation_results} reinforce this finding. Both OVI (with 6+ tokens) and the \texttt{TextEmb} baseline achieve higher average scores on T2I-CompBench++ than the specialized ECLIPSE prior. \texttt{TextEmb} achieves a score of 0.457, OVI with 6 tokens arrives at 0.450 and OVI with 9 tokens reaches 0.451, while ECLIPSE scores 0.410. Although ECLIPSE is superior in 3D spatial reasoning, \texttt{TextEmb} outperforms it in all other categories. This indicates a potential ``metric problem'' which we will analyze further in Section 5, where high benchmark scores do not necessarily correlate with better image quality.

\begin{table}[]
  \caption{\small Ablation study on T2I-CompBench++ (dim 512x512). We analyze the impact of varying the number of OVI steps, pseudo-tokens, and decoder diffusion steps. The performance of the unconstrained OVI method is highly sensitive to these parameters, with a higher number of tokens significantly improving the average score.}
  \label{tab:ablation_results}
  \centering
  \resizebox{1\textwidth}{!}{%
  \begin{tabular}{lccccccc}
    \toprule
    Method & Color & Shape & Texture & Spatial 2D & Spatial 3D & Non-Spatial & \textbf{Avg Score} \\
    \midrule
    ECLIPSE (50 diff) & 0.5633 & 0.4615 & 0.5582 & 0.1944 & \textbf{0.3709} & 0.3138 & 0.4104 \\
    \texttt{TextEmb} (50 diff) & \textbf{0.6029} & \textbf{0.5443} & \textbf{0.6953} & \textbf{0.2351} & 0.3403 & \textbf{0.3236} & \textbf{0.4569} \\
    \midrule
    OVI (1000 steps, 1 token, 50 diff) & 0.3019 & 0.3492 & 0.3618 & 0.0520 & 0.1829 & 0.2684 & 0.2527 \\
    OVI (600 steps, 1 token, 50 diff) & 0.2848 & 0.3309 & 0.3765 & 0.0612 & 0.1749 & 0.2718 & 0.2500 \\
    OVI (1200 steps, 1 token, 50 diff) & 0.2974 & 0.3476 & 0.4123 & 0.0465 & 0.1720 & 0.2797 & 0.2592 \\
    \midrule
    OVI (1000 steps, 3 tokens, 50 diff) & 0.5298 & 0.4865 & 0.6066 & 0.1474 & 0.2925 & 0.3184 & 0.3969 \\
    OVI (1000 steps, 6 tokens, 50 diff) & 0.5957 & 0.5526 & 0.6811 & 0.1984 & 0.3492 & 0.3225 & 0.4499 \\
    OVI (1000 steps, 9 tokens, 50 diff) & 0.5932 & 0.5403 & 0.6886 & 0.2215 & 0.3420 & 0.3222 & 0.4513 \\
    \midrule
    OVI (1000 steps, 1 token, 30 diff) & 0.2747 & 0.3344 & 0.3310 & 0.0437 & 0.1599 & 0.2649 & 0.2348 \\
    OVI (1000 steps, 1 token, 70 diff) & 0.2975 & 0.3437 & 0.3631 & 0.0552 & 0.1725 & 0.2715 & 0.2506 \\
    \bottomrule
  \end{tabular}
  }
\end{table}

\subsection{Improving Visual Quality with Priors and Constraints}

The success of ECLIPSE with a relatively low cosine similarity ($\approx$0.65) reveals a crucial insight: maximizing semantic similarity alone is not sufficient for generating high-quality images. The key is to produce an embedding that is not only semantically aligned but also lies within the manifold of realistic images. Our unconstrained OVI fails to do this. Therefore, to improve upon the strong but visually imperfect \texttt{TextEmb} and OVI baselines, we must introduce constraints that guide the optimization process towards this visual manifold.

\subsubsection{Mahalanobis-constrained OVI}
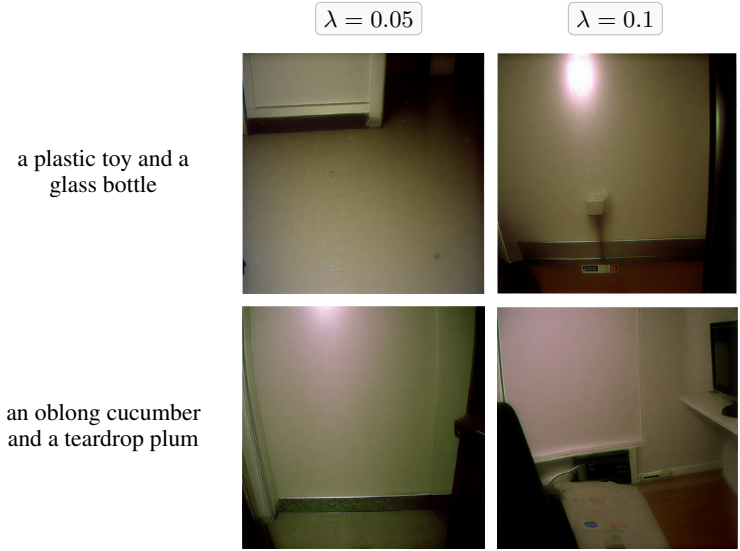
\begin{figure}[]
  % Rimuoviamo \centering per permettere lo spostamento manuale
  \hspace*{0.5cm} % <--- Modifica questo valore per spostare più o meno a sinistra
  \begin{minipage}{\textwidth} % Contenitore per mantenere l'allineamento interno
    
    % --- Stile dei titoli ---
    \tikzset{titlebox/.style={
      draw=gray!50, fill=gray!5, rounded corners=2pt, inner sep=3pt, font=\small\bfseries
    }}

    % --- Riga delle intestazioni Lambda ---
    \begin{minipage}[c]{0.20\textwidth} \hfill \end{minipage}
    \hspace{0pt} 
    \begin{minipage}[c]{0.28\textwidth}
      \centering
      \begin{tikzpicture}\node[titlebox] {$\lambda = 0.05$};\end{tikzpicture}
    \end{minipage}
    \hspace{-23pt} 
    \begin{minipage}[c]{0.28\textwidth}
      \centering
      \begin{tikzpicture}\node[titlebox] {$\lambda = 0.1$};\end{tikzpicture}
    \end{minipage}

    \vspace{6pt}

    % --- Prima riga di immagini ---
    \begin{minipage}[c]{0.20\textwidth}
      \flushright \small \centering a plastic toy and a glass bottle
    \end{minipage}
    \hspace{8pt} 
    \begin{minipage}[c]{0.23\textwidth}
      \centering
      \includegraphics[width=\textwidth]{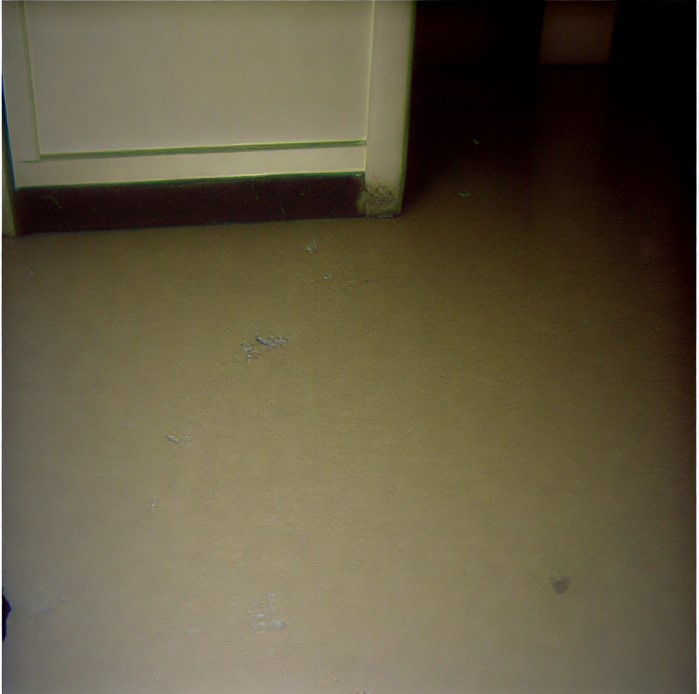}
    \end{minipage}
    \hspace{0pt} 
    \begin{minipage}[c]{0.227\textwidth}
      \centering
      \includegraphics[width=\textwidth]{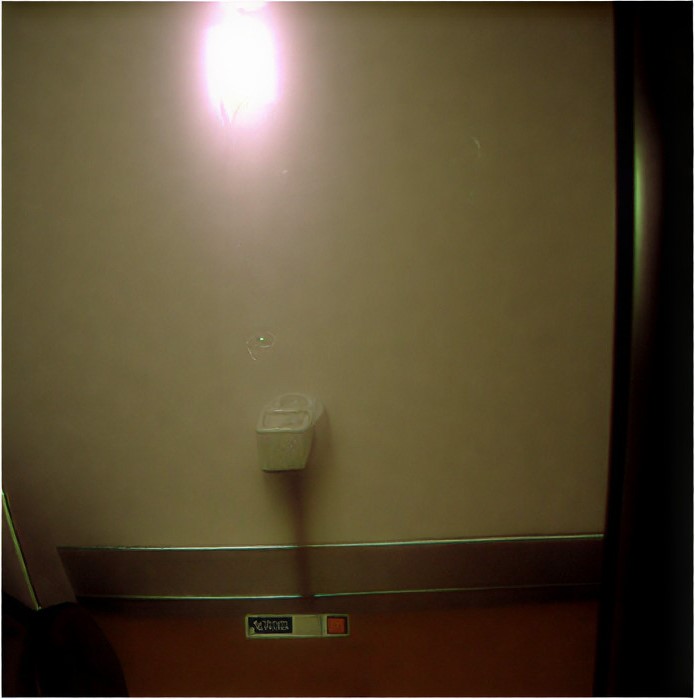}
    \end{minipage}

    \vspace{4pt} 

    % --- Seconda riga di immagini ---
    \begin{minipage}[c]{0.20\textwidth}
      \flushright \small \centering an oblong cucumber and a teardrop plum
    \end{minipage}
    \hspace{8pt}
    \begin{minipage}[c]{0.23\textwidth}
      \centering
      \includegraphics[width=\textwidth]{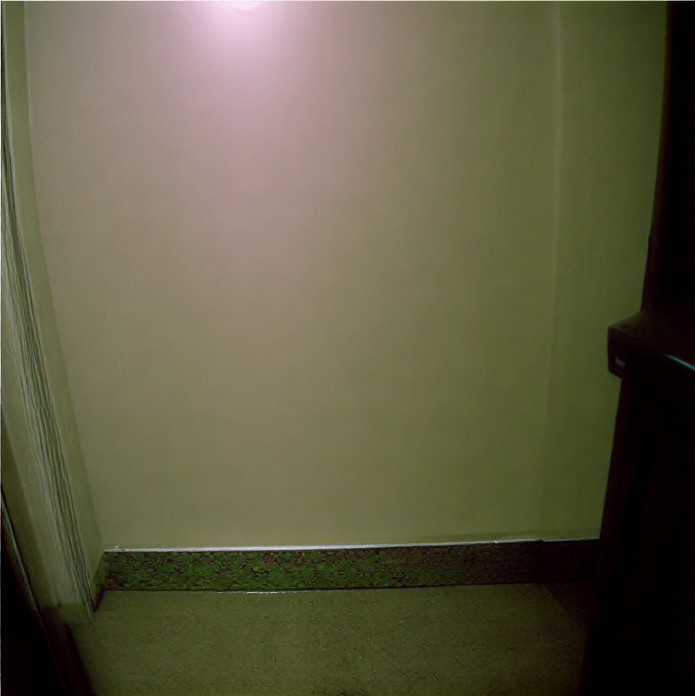}
    \end{minipage}
    \hspace{0pt}
    \begin{minipage}[c]{0.228\textwidth}
      \centering
      \includegraphics[width=\textwidth]{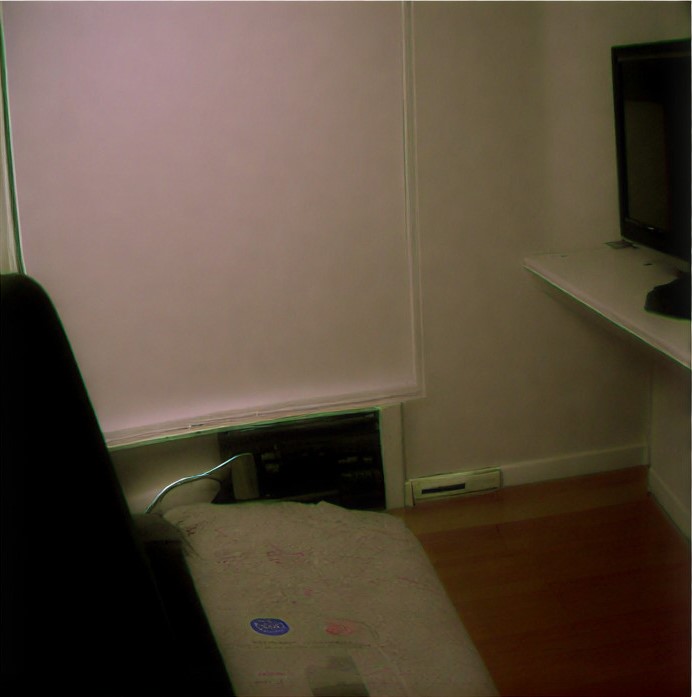}
    \end{minipage}
  \end{minipage}
  
  \caption{\small Mode collapse with high Mahalanobis constraints. Regardless of the input prompt, setting $\lambda_M \ge 0.05$ forces the generated image to converge towards a generic ``room'' representation, likely the mean of the COCO embedding space.}
  \label{fig:mahalanobis_collapse}
\end{figure}

We investigated the Mahalanobis constraint with various values for $\lambda_M$. We observed that without optimization, the Mahalanobis loss typically hovers around $75$. Our optimization successfully reduces this value significantly. However, we found that large values ($\lambda_M \ge 0.05$) caused the model to collapse, generating a generic ``room'' image regardless of the input prompt (likely the mean of the COCO dataset). This phenomenon is visually demonstrated in Figure \ref{fig:mahalanobis_collapse}.

\begin{figure}[!]
  \centering
  \includegraphics[width=\textwidth,height=1\textheight,keepaspectratio]{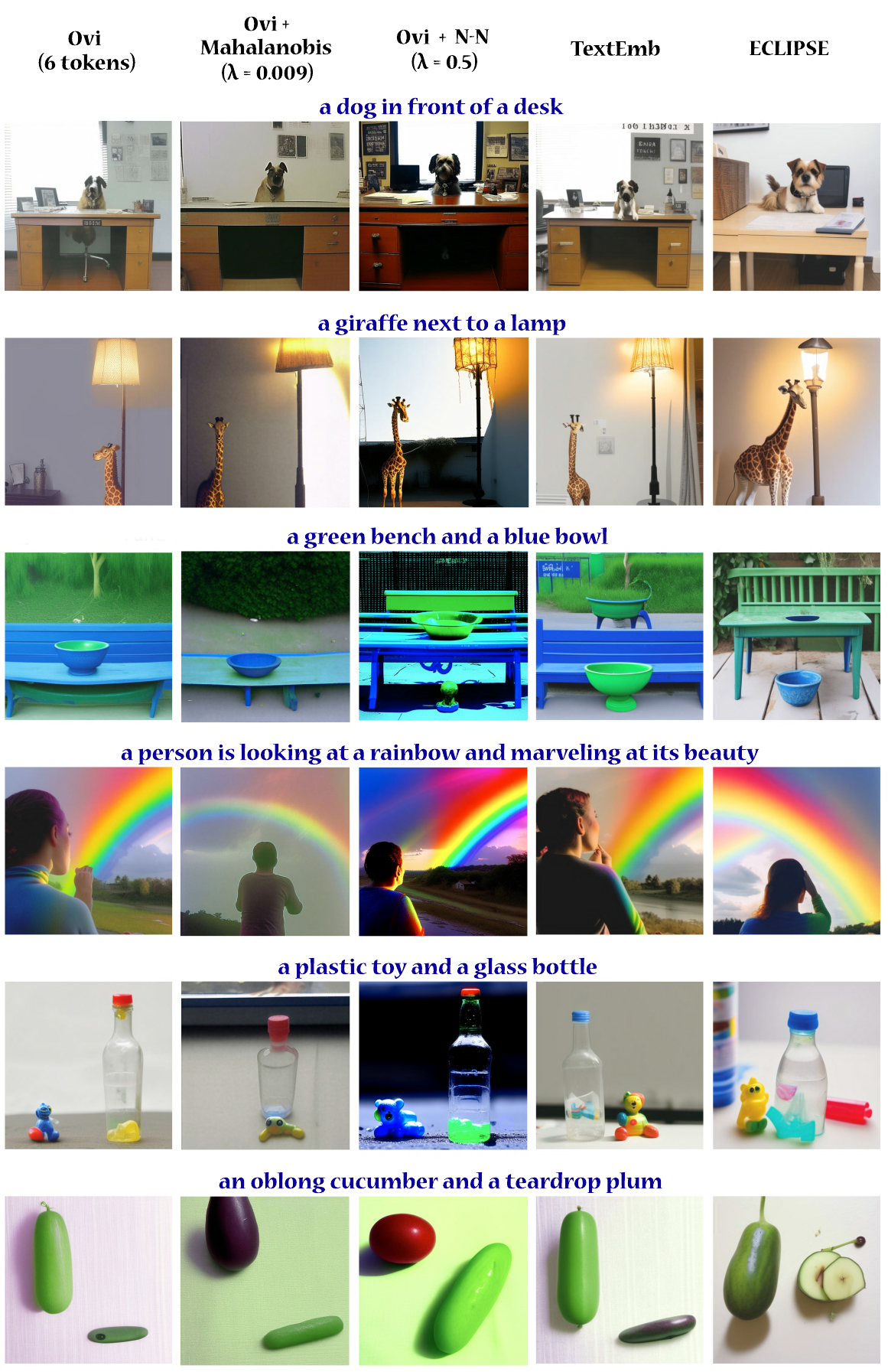}
  \caption{\small Qualitative comparison of constrained OVI methods. Both Mahalanobis and Nearest-Neighbor (NN) constraints improve visual quality over the base OVI and \texttt{TextEmb} baselines. The NN-constrained OVI produces images that are visually competitive with the trained ECLIPSE~\cite{Eclipse} prior.}
  \label{fig:constrained_ovi_qualitative}
\end{figure}

\begin{figure}[!]
  \centering

  % --- Prima riga: una sola immagine centrata ---
  \begin{subfigure}{0.48\textwidth}
    \centering
    \begin{tikzpicture}
      \node[titlebox] {Ovi (6 tokens)};
    \end{tikzpicture}
    \vspace{2pt} % Spazio tra riquadro e immagine
    \includegraphics[width=\textwidth]{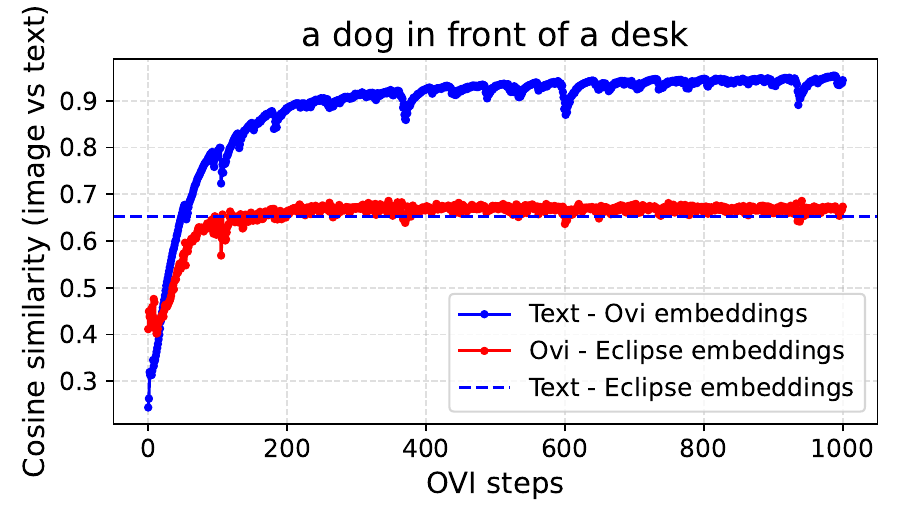}
  \end{subfigure}

  \vspace{1em} % Spazio verticale tra le righe

  % --- Seconda riga: due immagini affiancate ---
  \begin{subfigure}{0.48\textwidth}
    \centering
    \begin{tikzpicture}
      \node[titlebox] {Ovi + Mahalanobis ($\lambda = 0.009$)};
    \end{tikzpicture}
    \vspace{2pt}
    \includegraphics[width=\textwidth]{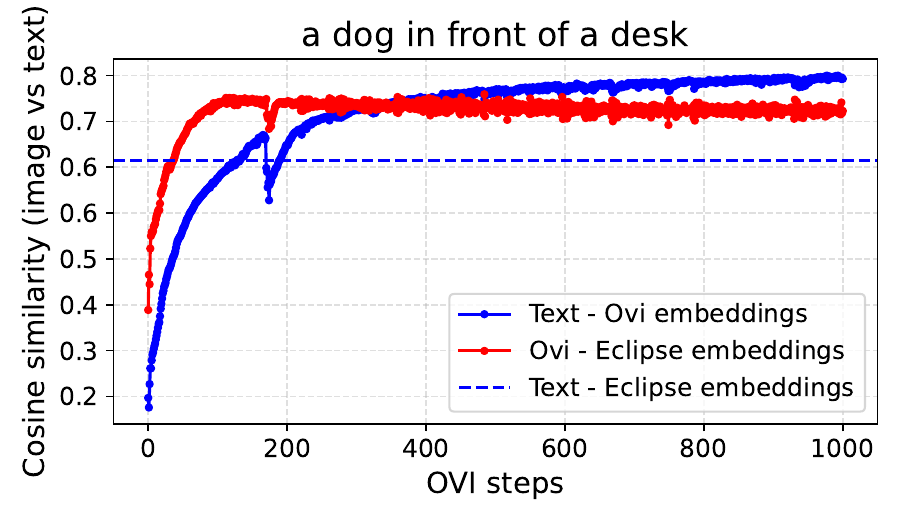}
  \end{subfigure}
  \hfill
  \begin{subfigure}{0.48\textwidth}
    \centering
    \begin{tikzpicture}
      \node[titlebox] {Ovi + N-N ($\lambda = 0.5$)};
    \end{tikzpicture}
    \vspace{2pt}
    \includegraphics[width=\textwidth]{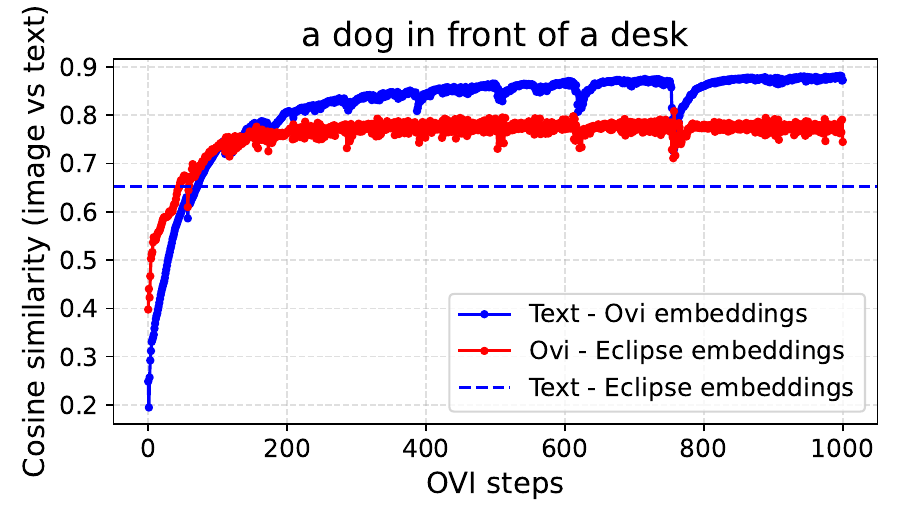}
  \end{subfigure}

  \caption{\small Evolution of cosine similarity metrics during the optimization process for different OVI configurations. The \textit{solid blue curve} tracks the similarity between the target text embedding and our optimized OVI embedding. The \textit{solid red curve} measures the similarity between our OVI embedding and the embedding generated by the ECLIPSE~\cite{Eclipse} prior. The \textit{dashed blue line} represents the baseline similarity between the text embedding and the ECLIPSE image embedding.
  \textbf{Top:} Unconstrained OVI (6 tokens) converges to the text embedding~($>0.9$). 
  \textbf{Bottom-Left:} Mahalanobis constraint ($\lambda_M=0.009$) restricts text similarity to $\approx 0.80$. 
  \textbf{Bottom-Right:} Nearest-Neighbor constraint ($\lambda_N=0.5$) stabilizes text similarity around $0.88$ while reaching a similarity of~$\approx 0.79$ with ECLIPSE, indicating strong convergence towards the trained prior’s manifold.}
  \label{fig:cosine_comparison}
\end{figure}

While $\lambda_M \approx 0.017$ offered a theoretical balance, we empirically found that a slightly lower value of $\lambda_M = 0.009$ yielded superior visual results (Figure \ref{fig:constrained_ovi_qualitative}). With this setting, the Mahalanobis loss converges to approximately 28. As visualized in Figure \ref{fig:cosine_comparison} (center), this constraint restricts the text-image cosine similarity to about 0.80, preventing it from reaching the near-perfect alignment ($>0.9$) seen in the unconstrained case. This indicates that the optimization is finding a compromise between semantic alignment and the statistical distribution of real images. Despite the visual improvement, the T2I-CompBench++ score dropped to 0.389 (Table \ref{tab:constrained}), further highlighting the trade-off between metric maximization and perceptual quality.

\subsubsection{Nearest-Neighbor Constrained OVI}
\begin{figure}[]
  \centering
  
  \tikzset{titlebox/.style={
    draw=gray!50, 
    fill=gray!5, 
    rounded corners=2pt, 
    inner sep=3pt, 
    font=\fontsize{8}{10}\selectfont
  }}

  % --- PRIMA RIGA ---
  \begin{subfigure}[b]{0.32\textwidth}
    \centering
    % Area del titolo con altezza fissa per garantire la separazione
    \begin{minipage}[t][1.2cm][c]{\textwidth}
      \centering
      \begin{tikzpicture}
        \node[titlebox, text width=2.8cm, align=center] {a green bench and a blue bowl};
      \end{tikzpicture}
    \end{minipage}
    \vspace{4pt} % Spazio extra tra l'area del titolo e l'immagine
    \includegraphics[width=\textwidth, height=2.5cm, keepaspectratio]{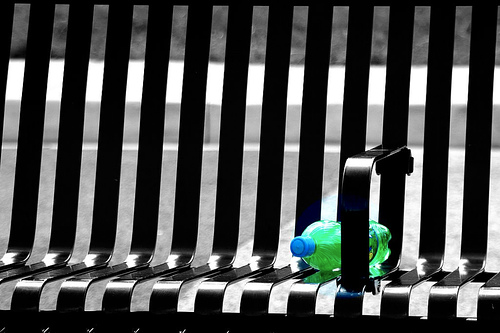}
  \end{subfigure}
  \hfill
  \begin{subfigure}[b]{0.32\textwidth}
    \centering
    \begin{minipage}[t][1.2cm][c]{\textwidth}
      \centering
      \begin{tikzpicture}
        \node[titlebox, text width=2.8cm, align=center] {a giraffe next to a lamp};
      \end{tikzpicture}
    \end{minipage}
    \vspace{4pt}
    \includegraphics[width=\textwidth, height=3.6cm, keepaspectratio]{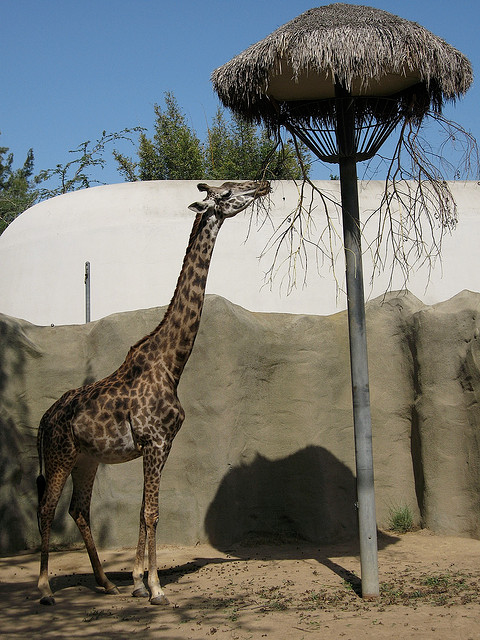}
  \end{subfigure}
  \hfill
  \begin{subfigure}[b]{0.32\textwidth}
    \centering
    \begin{minipage}[t][1.2cm][c]{\textwidth}
      \centering
      \begin{tikzpicture}
        \node[titlebox, text width=2.8cm, align=center] {a plastic toy and a glass bottle};
      \end{tikzpicture}
    \end{minipage}
    \vspace{4pt}
    \includegraphics[width=\textwidth, height=2.5cm, keepaspectratio]{coco1.png}
  \end{subfigure}

  \vspace{7pt} 

  % --- SECONDA RIGA ---
  \begin{subfigure}[b]{0.32\textwidth}
    \centering
    \begin{minipage}[t][1.2cm][c]{\textwidth}
      \centering
      \begin{tikzpicture}
        \node[titlebox, text width=2.8cm, align=center] {a dog in front of a desk};
      \end{tikzpicture}
    \end{minipage}
    \vspace{4pt}
    \includegraphics[width=\textwidth, height=3.6cm, keepaspectratio]{coco3.png}
  \end{subfigure}
  \hfill
  \begin{subfigure}[b]{0.32\textwidth}
    \centering
    \begin{minipage}[t][1.2cm][c]{\textwidth}
      \centering
      \begin{tikzpicture}
        \node[titlebox, text width=2.8cm, align=center] {an oblong cucumber and a teardrop plum};
      \end{tikzpicture}
    \end{minipage}
    \vspace{4pt}
    \includegraphics[width=\textwidth, height=2.5cm, keepaspectratio]{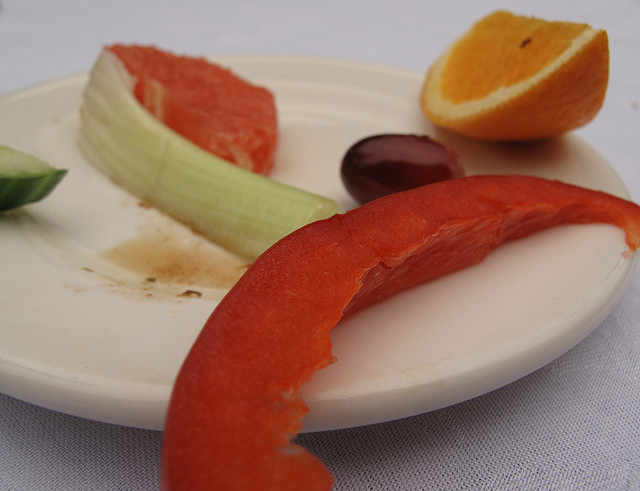}
  \end{subfigure}
  \hfill
  \begin{subfigure}[b]{0.32\textwidth}
    \centering
    \begin{minipage}[t][1.2cm][c]{\textwidth}
      \centering
      \begin{tikzpicture}
        \node[titlebox, text width=2.8cm, align=center] {a person is looking at a rainbow and \hspace{2cm}marveling at its beauty};
      \end{tikzpicture}
    \end{minipage}
    \vspace{4pt}
    \includegraphics[width=\textwidth, height=2.7cm, keepaspectratio]{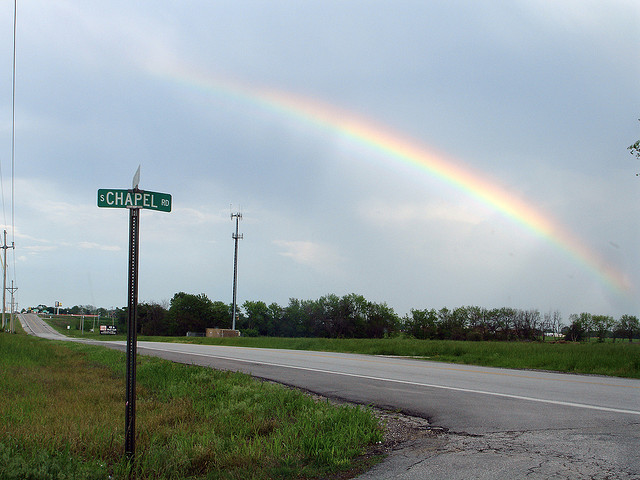}
  \end{subfigure}
  
  \vspace{10pt}
  
  \caption{\small Nearest neighbors retrieved from the MS-COCO~\cite{lin2014microsoft} dataset for six different text prompts. These real images serve as the reference ``anchors'' ($z_{closest}$) for our Nearest-Neighbor Constrained OVI, guiding the optimization towards realistic visual compositions found in the dataset.}
  \label{fig:coco_neighbors}
\end{figure}

To illustrate the nature of this guidance, Figure \ref{fig:coco_neighbors} displays the nearest neighbor images retrieved from the MS-COCO dataset. Typically, the cosine distance (neighbor loss) between a generated unconstrained embedding and its nearest neighbor is around $0.55$. With aggressive optimization ($\lambda_N=4$), we can drive this distance down to $0.026$, but this overrides the prompt semantics.

The Nearest-Neighbor constraint proved most effective with a balanced $\lambda_N = 0.5$. In this configuration, the neighbor loss converges to approximately 0.28. This approach generates images of significantly higher visual quality, often achieving near-ECLIPSE performance. (Figure \ref{fig:constrained_ovi_qualitative}). As shown in Figure \ref{fig:cosine_comparison} (right), the optimization stabilizes the text cosine similarity around 0.88. This is lower than the unconstrained baseline but higher than the Mahalanobis variant, suggesting that anchoring to a specific real image allows for better semantic retention than a global distributional constraint. In addition, this setting achieves a similarity of approximately 0.79 with the ECLIPSE embedding, one of the highest recorded values, confirming that we are effectively approaching the visual manifold learned by the trained prior. On the T2I-CompBench++ benchmark, the method achieves an average score of 0.415, outperforming ECLIPSE (0.410) as detailed in Table \ref{tab:constrained}.

\begin{table}[]
  \caption{\small T2I-CompBench++ main results (dim 512x512). We compare our best constrained OVI methods against the trained ECLIPSE prior and the strong \texttt{TextEmb} baseline. The NN-constrained OVI improves visual quality while also outperforming ECLIPSE.}
  \label{tab:constrained}
  \centering
  \resizebox{\textwidth}{!}{%
  \begin{tabular}{lccccccc}
    \toprule
    Method & Color & Shape & Texture & Spatial 2D & Spatial 3D & Non-Spatial & \textbf{Avg Score} \\
    \midrule
    ECLIPSE (trained) & 0.563 & 0.461 & 0.558 & 0.194 & \textbf{0.371} & 0.314 & 0.410 \\
    \texttt{TextEmb} (training-free) & \textbf{0.603} & \textbf{0.544} & \textbf{0.695} & \textbf{0.235} & 0.340 & \textbf{0.324} & \textbf{0.457} \\
    \midrule
    OVI (base, 6 tokens) & 0.596 & 0.553 & 0.681 & 0.198 & 0.349 & 0.322 & 0.450 \\
    OVI + Mahalanobis ($\lambda=0.009$) & 0.547 & 0.473 & 0.636 & 0.103 & 0.247 & 0.327 & 0.389 \\
    OVI + Nearest-Neighbor ($\lambda=0.5$) & 0.556 & 0.482 & 0.602 & 0.183 & 0.342 & 0.326 & 0.415 \\
    \bottomrule
  \end{tabular}
  }
\end{table}

%======================================================================
\section{Discussion and Limitations}

\textbf{The Metric Problem: High Scores vs. Visual Quality.}
Our most striking finding is the discrepancy between the quantitative scores from T2I-CompBench++ and the perceived visual quality of the generated images. The \texttt{TextEmb} baseline, despite its lower visual fidelity, consistently outperforms a strong trained prior like ECLIPSE~\cite{Eclipse}. This is not because the metrics are simply ``coupled with raw CLIP similarity,'' but due to a more nuanced reason revealed by our analysis.

The benchmark metrics (such as those based on Disentangled BLIP-VQA~\cite{li2022blip}) often verify the presence of atomic attributes (e.g., “is the book red?” or “is there a car?”) and the raw text embedding is the purest representation of these attributes, hence gives the highest score.
Instead, a prior projects this representation into the natural image manifold.
% , introducing coherence, lighting, depth, and realism, which these metrics overlook. 
In this sense, the text embedding is well-suited for satisfying discrete directives, while the value of a true prior (trained, like ECLIPSE, or optimized, like our constrained OVI) lies in integrating these factors into a unified and perceptually rich composition. This highlights the need for holistic evaluation metrics that balance atomic correctness with overall visual quality.

\textbf{Limitations.}
Although promising, our OVI approach has limitations. Its primary drawback is inference speed. The iterative optimization process (1000 steps) is orders of magnitude slower than a single forward pass through a trained prior like ECLIPSE. Secondly, our constrained methods depend on an external dataset (MS-COCO). While deriving statistics for the Mahalanobis constraint is a one-time pre-computation, the Nearest-Neighbor constraint introduces additional computational overhead during inference. Specifically, for every new prompt, the method requires computing the cosine similarity between the input text embedding and the entire reference dataset to identify the closest match.

%======================================================================
\section{Conclusions and Future Work}

In this work, we investigated Optimization-based Visual Inversion (OVI) as a training-free diffusion prior in the context of text-to-image generation. Our experiments indicate that this approach is a viable alternative to traditional expensive trained priors and, in its unconstrained form, helped expose a critical weakness in current compositional T2I benchmarks. Directly feeding the text embedding into the decoder often outperforms expensive state-of-the-art priors in the evaluated benchmark. Additionally, we show that by incorporating Mahalanobis and Nearest-Neighbor constraints, we can steer the OVI-based optimization process to produce higher visual fidelity images. Our Nearest-Neighbor constrained OVI, in particular, achieves a better quantitative score and competitive visual quality compared to the leading ECLIPSE prior, all without requiring any prior training and utilizing a reference set orders of magnitude smaller ($\approx$0.8\% of the latter's training data).

For future work, we see several promising directions. The Nearest-Neighbor constraint could be enhanced by using a K-Nearest Neighbors approach, averaging the embeddings of several close neighbors to provide a more robust target. 
% Furthermore, its performance is contingent on the size and diversity of the reference dataset; expanding this dataset from 40k images to millions would likely improve results by providing more relevant targets for optimization. 
A Gaussian Mixture Model~\cite{bishop2006pattern} could also be used to model the distribution of image embeddings more accurately than a single Gaussian. Most importantly, our work underscores the need for better evaluation metrics that capture the full spectrum of image quality, including photorealism, coherence, and compositional correctness, beyond simple semantic similarity.

%======================================================================
\bibliographystyle{plain}
\bibliography{references}

\end{document}